\documentclass[letterpaper]{article} 
\usepackage{aaai2027}  
\nocopyright
\usepackage{amsfonts}
\usepackage{amsmath}
\usepackage[hyphens]{url}  
\usepackage{graphicx} 
\usepackage{natbib}  
\usepackage{caption} 
\usepackage{algorithm}
\usepackage{algorithmic}

\usepackage{newfloat}
\usepackage{listings}
\DeclareCaptionStyle{ruled}{labelfont=normalfont,labelsep=colon,strut=off} 
\floatstyle{ruled}
\newfloat{listing}{tb}{lst}{}
\floatname{listing}{Listing}

\usepackage{booktabs}

\title{DA-NBV: A Direction-Aware Next-Best-View Planner for Efficient 3D Reconstruction of Ships at Sea}

\author{
Jiaming Chen\textsuperscript{\rm 1}\equalcontrib,
Juntao Yang\textsuperscript{\rm 1}\equalcontrib,
Zhentao Zou\textsuperscript{\rm 2},
Qi Ming\textsuperscript{\rm 3},\\
Yi Yu\textsuperscript{\rm 4},
Zhihang Zhong\textsuperscript{\rm 2},
Xue Yang\textsuperscript{\rm 2},
Xue Jiang\textsuperscript{\rm 2},
Yue Zhou\textsuperscript{\rm 1}\corresponding
}

\affiliations{
\textsuperscript{\rm 1}East China Normal University\\
\textsuperscript{\rm 2}Shanghai Jiao Tong University\\
\textsuperscript{\rm 3}Beijing University of Technology\\
\textsuperscript{\rm 4}The Ohio State University\\
}

\begin{document}

\maketitle

\begin{abstract}
Accurate 3D reconstruction of ships at sea is important for maritime supervision, damage assessment, and autonomous maritime operations. Although 3D reconstruction has advanced considerably, high-quality data acquisition still largely relies on manually designed trajectories or skilled operators, resulting in high costs and limited scalability. Next-best-view (NBV) planning automates this process by selecting subsequent viewpoints based on the current state.
However, existing NBV policies mainly model spatial occupancy while overlooking directional observation history. This limitation is particularly problematic for ships: their complex superstructures and severe self-occlusions require observations from multiple viewpoints, and insufficient directional coverage often yields incomplete reconstructions. These challenges are further amplified at sea, where wave-induced heave, roll, and pitch continuously alter the ship's pose and surface visibility. Meanwhile, wind disturbances and limited onboard power impose stricter requirements on scanning efficiency.
To address these challenges, we propose DA-NBV, a direction-aware NBV policy that augments the conventional occupancy state with directional observation statistics. We introduce a learnable Position Advantage Field (PAF) that uses directional information to guide viewpoint selection. The policy further adopts a locally constrained action space and a nonlinear coverage-shaping reward to improve scanning efficiency. We also develop the ship-oriented SeaShip-3D dataset and a configurable sea-state simulation environment. Experiments under varying heave, roll, and pitch conditions show that DA-NBV improves reconstruction completeness by approximately 3 percentage points and reduces Chamfer distance by 43\% while achieving higher path efficiency.

\end{abstract}

\section{Introduction}
Three-dimensional reconstruction of ships at sea using unmanned aerial vehicles (UAVs) is important for vessel maintenance, maritime supervision, and digital asset management~\cite{shah2021condition,liu2022wind}. Accurate models support structural measurement, corrosion and damage assessment, vessel identification, modification detection, cargo load assessment, and port management, while reducing the risks associated with manual inspection at height or near water. They also provide a geometric foundation for digital twins, simulation, and lifecycle records. Moreover, UAVs can reconstruct vessels while they are underway without requiring berthing or disrupting voyage schedules. Operating in such dynamic scenes, however, requires policies that can adapt to vessel motion.

Although advances in 3D reconstruction have simplified image-based digitization, image acquisition remains costly and inefficient because it is still largely manual. To streamline real-world digitization, recent studies have explored next-best-view (NBV) policies for automatic image acquisition using drones~\cite{maboudi2023review,peralta2020scanrl,chen2024gennbv}. Heuristic NBV policies can achieve high coverage in specific scenes but generalize poorly~\cite{connolly1985nbv,pito1999solution,isler2016informationgain,delmerico2018comparison}. RL-based methods improve generalization while maintaining high coverage~\cite{peralta2020scanrl,chen2024gennbv}. GenNBV additionally encodes geometry with a probabilistic 3D grid~\cite{chen2024gennbv}.

\begin{figure*}[t]
    \centering
    \includegraphics[width=\textwidth]{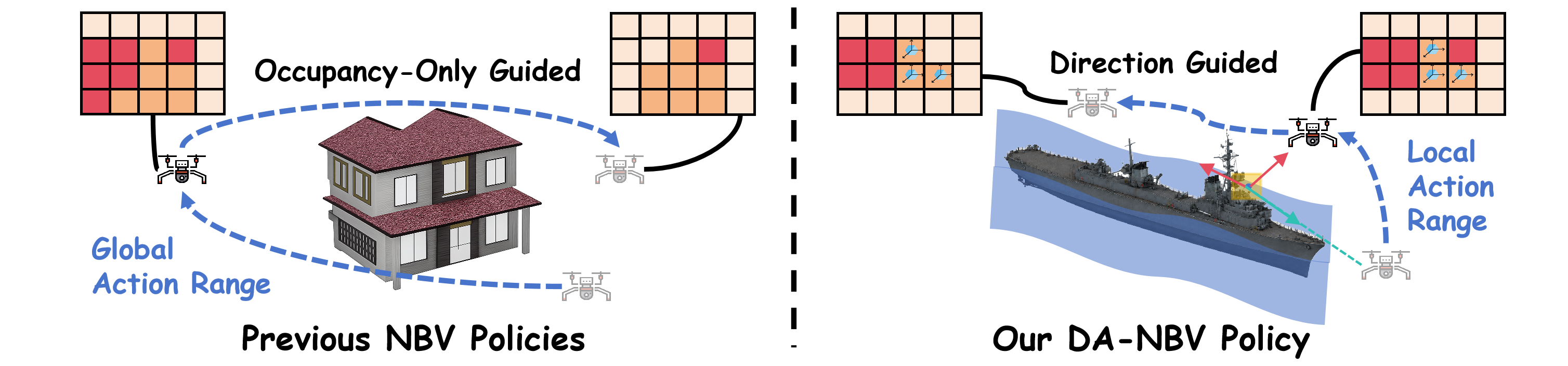}
    \caption{
Comparison of conventional and DA-NBV scanning strategies. Conventional NBV methods are primarily designed for static objects and rely on occupancy-based representations that distinguish unknown, free, and occupied space. Their global action spaces can also lead to redundant observations and unnecessarily long trajectories. In contrast, DA-NBV is robust to dynamic targets, uses missing-direction information to acquire complementary views of the same region, and adopts a local action space to generate more efficient scanning trajectories.
}
    \label{fig:comparison}
\end{figure*}

As illustrated in Fig.~\ref{fig:comparison}, a fundamental limitation of existing occupancy-based NBV policies is that they record whether a voxel has been observed, but not the diversity of directions from which it has been viewed. Once a voxel becomes visible, it is often treated as sufficiently covered, even when its geometry remains poorly captured. This limitation is particularly significant for ships, whose complex superstructures and extensive self-occlusions require observations from multiple complementary directions. Without directional awareness, the policy may repeatedly select similar viewpoints or viewpoints with limited surface visibility, resulting in redundant observations, persistent blind spots, and inefficient trajectories. These challenges become even more severe in maritime environments, where wave-induced heave, roll, and pitch continuously change the pose and surface visibility of the ship. Meanwhile, wind disturbances increase UAV energy consumption, and limited opportunities for landing or recharging may leave the UAV with only one chance to complete the scan. Reconstructing ships at sea therefore requires efficient trajectories, high reconstruction quality, and robust policies~\cite{liu2022wind}.

To meet these requirements and extend NBV planning to dynamic reconstruction scenarios, we propose DA-NBV, a Direction-Aware Next-Best-View Planner for Efficient 3D Reconstruction of Ships at Sea.

The core idea is that geometrically complex regions deserve observations from a more diverse set of directions. Although one or two observations may suffice for smooth hull surfaces, complex structures such as ship superstructures require complementary views from multiple directions to support high-quality reconstruction.

We extend the occupancy-based planning pipeline through three coupled components: a direction-aware reconstruction state with a learnable
Position Advantage Field (PAF), a locally constrained autoregressive action policy, and nonlinear coverage shaping. Together, these components encourage complementary observations while simplifying local search and reducing unnecessary flight.

We also construct the ship-oriented SeaShip-3D dataset and use it to build a physics-based maritime scanning environment for training and evaluation. Experiments demonstrate that DA-NBV achieves greater path efficiency, fewer blind spots, and improved reconstruction quality compared with existing NBV planning methods.

\section{Related Work}

Next-best-view (NBV) planning selects the next sensor viewpoint according to the current reconstruction state~\cite{bajcsy1988active,scott2003viewplanning,vasquez2009viewplanning}. Traditional methods typically generate a set of candidate viewpoints and rank them using handcrafted utility functions based on information gain, unknown-space coverage, visibility, or motion cost~\cite{connolly1985nbv,pito1999solution,kriegel2011surface,doumanoglou2016recovering,isler2016informationgain,delmerico2018comparison}. Although these methods perform well in specific environments, their effectiveness strongly depends on manually designed functions. Consequently, they often exhibit limited generalization.

Learning-based NBV methods can be broadly divided into scene-specific online optimization methods and policies pretrained offline~\cite{hepp2018learntoscore,zeng2020pcnbv,jin2025activegs,xu2025hgs,tao2025rtguide}. NeRF-based active reconstruction methods usually optimize a scene-specific neural representation while selecting subsequent observations~\cite{mildenhall2020nerf,pan2022activenerf,lee2022uncertainty,yan2023activeimplicit,xue2024nvf,xiao2024nerfdirector,jiang2024fisherrf}. For example, ActiveRMAP alternates between radiance-field reconstruction and path optimization during scanning~\cite{zhan2022activermap}. Although these methods can directly exploit the current neural scene representation, they require repeated scene-level optimization and rendering, resulting in substantial computational overhead and limiting their suitability for real-time deployment. In contrast, reinforcement-learning-based methods train reusable policies in simulation and generate the next viewpoint through feedforward inference during deployment. Scan-RL demonstrates the feasibility of learning a scanning policy from image histories~\cite{peralta2020scanrl}. GenNBV further combines probabilistic 3D geometry, image features, and action history, enabling stronger cross-dataset generalization~\cite{chen2024gennbv}.

The effectiveness of NBV policies depends strongly on how the evolving reconstruction state is represented~\cite{hornung2013octomap,guedon2022scone,peralta2020scanrl,chen2024gennbv}. NBV-Net and GenNBV use probabilistic occupancy grids to represent whether individual voxels have been observed~\cite{mendoza2020supervised,chen2024gennbv}. Hestia further models the observation states of the six voxel faces, providing a finer-grained representation of reconstruction progress~\cite{lu2026hestia}. Nevertheless, these representations primarily characterize spatial coverage. Consequently, they may not adequately distinguish repeated observations from similar viewpoints from complementary observations acquired from previously unexplored directions. Therefore, high surface coverage does not always guarantee accurate reconstruction, particularly for self-occluded regions and fine geometric structures~\cite{frahm2025vinnbv}.

Existing methods also differ substantially in their action-space designs. Traditional approaches usually select viewpoints from limited predefined candidate sets~\cite{connolly1985nbv,pito1999solution}, whereas recent methods allow viewpoint selection over larger and more flexible action spaces~\cite{chen2024gennbv,lu2026hestia}. However, expanding the global action space makes policy optimization more difficult and often results in redundant back-and-forth trajectories. Moreover, selecting viewpoints according to one-step utility does not directly optimize long-term path efficiency~\cite{schmid2012viewplanning,roberts2017submodular,hepp2018plan3d,maboudi2023review}. In contrast, DA-NBV explicitly maintains voxel-level historical observation directions and constructs a direction-aware Position Advantage Field to identify candidate positions that provide complementary observations. Its locally constrained autoregressive policy first selects a relative position and then predicts the camera orientation and termination action. This factorization reduces the single-step search space while allowing the UAV to progressively explore a larger free space through successive local movements.


\section{Method}
\label{sec:method}

\begin{figure*}[t]
    \centering
    \includegraphics[width=\textwidth]{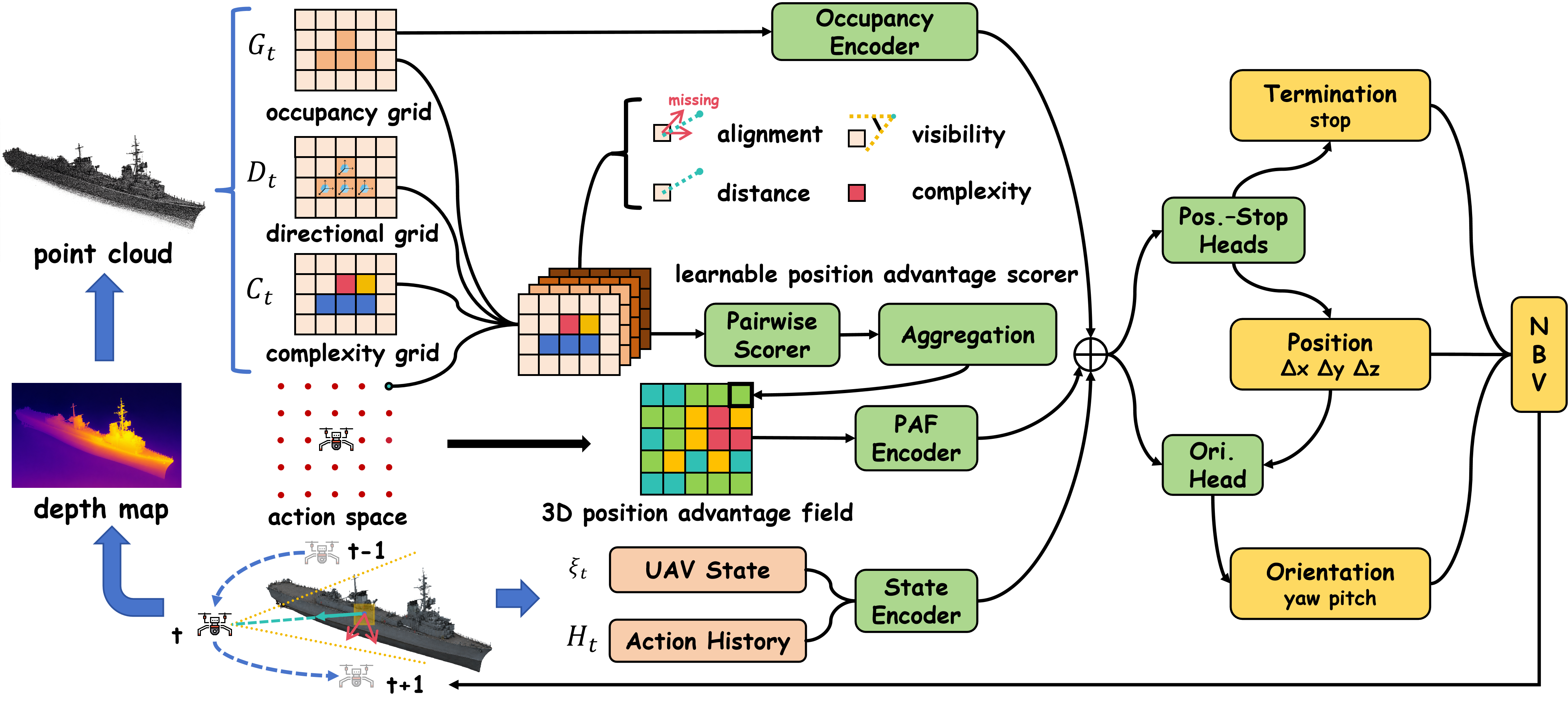}
    \caption{
    Overview of DA-NBV.
    Motion-compensated depth observations update the occupancy grid $G_t$, voxel-wise direction-vector grid $D_t$, and geometric complexity grid $C_t$.
    The Learnable Position Advantage Scorer (LPAS) evaluates voxel--candidate-position pairs using viewing-direction alignment, visibility, distance, and geometric complexity, and aggregates their utilities into the PAF $\mathbf U_t$.
    The encoded occupancy, PAF, UAV state, and action history are fused to predict position, orientation, and termination.
    }
    \label{fig:framework}
\end{figure*}

\subsection{DA-NBV Framework}
\label{sec:framework}

We formulate DA-NBV planning as a Markov decision process (MDP)
$\mathcal M=(\mathcal S,\mathcal A,\mathcal P,\mathcal R,\gamma)$,
where $\mathcal S$ denotes the reconstruction-state space,
$\mathcal A$ denotes the viewpoint-action space,
$\mathcal P$ denotes the state transition induced by UAV motion,
ship motion, and new depth observations,
$\mathcal R$ denotes the scanning reward,
and $\gamma$ denotes the discount factor.
Although the complete ship geometry is not directly observable,
the accumulated reconstruction grids and action history provide
a compact state approximation for sequential decision making.

As shown in Fig.~\ref{fig:framework}, DA-NBV forms a closed-loop
process from depth-based state construction to direction-aware
position scoring and action prediction. At decision step $t$, the
policy receives the state representation
$s_t=\{G_t,D_t,C_t,\xi_t,H_t\}$, where $G_t$ encodes occupied, free, and unknown space~\cite{thrun2005probabilistic};
$D_t$ encodes voxel-wise viewing-direction vectors and their
accumulated observation states;
$C_t$ describes local geometric complexity;
$\xi_t$ denotes the current UAV pose and motion state;
and $H_t$ contains the recently executed actions.

The current depth image is back-projected into a point cloud $P_t$. Since wave-induced heave, roll, pitch, and translation continuously alter the ship's pose, directly integrating observations in a static world coordinate system would introduce ghosting and geometric distortion. We therefore maintain a unified ship-centered reconstruction map. The first observation initializes the accumulated map $M_0$. At each subsequent step, iterative closest point (ICP) estimates the relative transformation $T_t$ between the current point cloud and the historical map. The aligned point cloud $\widetilde P_t=T_tP_t$ is then fused with $M_{t-1}$ to obtain the updated map $M_t$~\cite{besl1992icp}. The aligned observations are used to update $G_t$, $D_t$, and $C_t$, ensuring that the policy operates on a spatially consistent reconstruction state. Given $s_t$, the policy selects an action $a_t\in\mathcal A$. After the action is executed, the UAV acquires a new observation. The environment then transitions to a state $s_{t+1}$ sampled from $\mathcal P(\cdot\mid s_t,a_t)$, and the agent receives reward $r_t=\mathcal R(s_t,a_t,s_{t+1})$.

The PAF derives structured candidate-position guidance from $G_t$, $D_t$, and $C_t$. The locally constrained autoregressive policy defines the action distribution $\pi(a_t\mid s_t)$, while the reward function evaluates the resulting directional coverage, surface coverage, and path efficiency.

Following the three encoding branches in Fig.~\ref{fig:framework}, the occupancy encoder processes $G_t$, the PAF encoder processes $\mathbf U_t$, and the state encoder processes $\xi_t$ and $H_t$. Their outputs are concatenated as $h_t=\operatorname{Concat}(F_{\mathrm{occ}}(G_t),F_{\mathrm{PAF}}(\mathbf U_t),F_{\mathrm{state}}(\xi_t,H_t))$, where $h_t$ is used by the subsequent action branches.

\subsection{Position Advantage Field (PAF)}
\label{sec:paf}

The MDP state $s_t$ must capture not only which regions have already been observed but also which viewing directions remain underexplored and how important each region is for reconstruction. DA-NBV therefore represents each valid observation as a unit direction vector and accumulates these vectors at the voxel level. The resulting vector-based directional state distinguishes between redundant observations from similar directions and complementary observations from new directions. In addition, we estimate each region's geometric complexity online to quantify its importance for reconstruction. Using this state, DA-NBV constructs a 3D PAF from directional incompleteness, local geometric complexity, and candidate-position visibility. The resulting field $\mathbf U_t$ provides state-dependent guidance.

For each visible voxel, DA-NBV represents the corresponding observation as a unit viewing-direction vector. To store these vectors in a fixed-size voxel representation, we discretize the unit sphere into $N_d$ approximately uniformly distributed directional bins $\{b_j\}_{j=1}^{N_d}$. Each bin represents an angular range, so small changes caused by ship oscillation can still activate the same bin, improving robustness while preserving directional guidance. For a voxel $v$ centered at $x_v$ and a camera located at $p$, the voxel-centered viewing-direction vector is defined as
\begin{equation}
    d(v,p)=
    \frac{p-x_v}
    {\left\|p-x_v\right\|}.
    \label{eq:view-direction}
\end{equation}
When voxel $v$ is visible in the current depth observation, we identify
the directional bin whose vector $b_j$ is most aligned with $d(v,p)$
and set $B_t(v,j)=1$. Here, $B_t(v,j)\in\{0,1\}$ records whether voxel
$v$ has been physically observed from a direction aligned with $b_j$ by
time step $t$.

At each decision step, DA-NBV estimates voxel-wise geometric complexity online from the accumulated depth observations. We incrementally maintain local point statistics and compute a PCA-based descriptor $\hat{\mathbf c}_t(v)$ comprising linearity, scattering, and curvature~\cite{weinmann2015semantic}. This observation-derived descriptor provides an online estimate of voxel importance for the LPAS and PAF construction. 

Visibility estimation prevents occluded voxels from incorrectly increasing the advantage of a candidate position. Since exact ray tracing over all voxel--candidate pairs is computationally expensive, we approximate the transmittance of each occupied voxel using the axis-aligned bounding box of its accumulated point cloud. Let
$\Delta_k(u)=x_{u,k}^{\max}-x_{u,k}^{\min}$ denote its extent along axis $k$. The transmittance along axis $k\in\{x,y,z\}$ is estimated as
\begin{equation}
\tau_k(u)=
1-
\frac{\Delta_i(u)\Delta_j(u)}{A_{ij}},
\qquad
\{i,j,k\}=\{x,y,z\},
\label{eq:axis-transmittance}
\end{equation}
where $A_{ij}$ is the corresponding voxel-face area perpendicular to axis $k$. For an arbitrary viewing direction $\mathbf d$, the directional transmittance is obtained by cosine-weighted interpolation:
\begin{equation}
\tau(u,\mathbf d)=
\frac{\sum_{k\in\{x,y,z\}}|d_k|\tau_k(u)}
{\sum_{k\in\{x,y,z\}}|d_k|+\varepsilon}.
\label{eq:directional-transmittance}
\end{equation}
Given a target voxel $v$ and candidate position $q$, we sample the occupied voxels $\mathcal R(v,q)$ along the line segment connecting $v$ and $q$ and estimate the overall visibility as
\begin{equation}
\operatorname{Vis}(v,q)=
\prod_{u\in\mathcal R(v,q)}
\tau\left(
u,
d(v,q)
\right).
\label{eq:path-visibility}
\end{equation}
This conservative approximation may overestimate occlusion, but it efficiently suppresses contributions from invisible voxels without expensive exact ray tracing.

For a candidate position $q$, the unit direction vector from the voxel center toward the candidate is denoted by $d(v,q)$. We evaluate whether the candidate provides a missing viewing direction by comparing $d(v,q)$ with the unobserved direction vectors $\{b_j\}$:
\begin{equation}
    \rho_t(v,q)=
    \max_{\substack{
        j:B_t(v,j)=0
    }}
        d(v,q)^\top b_j
.
    \label{eq:direction-alignment}
\end{equation}
A high value of $\rho_t(v,q)$ indicates that the candidate viewing-direction vector is strongly aligned with a direction that has not yet been observed.

Within the local action region, we uniformly sample a structured set of representative positions $\mathcal Q_t$. As illustrated in Fig.~\ref{fig:framework}, the Learnable Position Advantage Scorer (LPAS) evaluates each voxel--candidate pair $(v,q)$ using its voxel-to-candidate distance $\delta_t(v,q)$, visibility $\operatorname{Vis}(v,q)$, missing-direction alignment $\rho_t(v,q)$, and the online geometric descriptor $\hat{\mathbf c}_t(v)$. The latter comprises the linearity, scattering, and curvature of voxel $v$.

For each candidate position, the LPAS estimates a pairwise utility for every active voxel from the corresponding voxel--candidate features. These voxel-wise utilities are summed to obtain a single advantage score for the candidate. This aggregation favors not only positions that are highly informative for individual voxels but also positions that can jointly observe multiple active regions from previously unobserved directions. The LPAS and the policy network are jointly optimized in an end-to-end manner.

The candidate advantages are then placed at their corresponding coordinates in the local 3D candidate space to construct the PAF $\mathbf U_t$. This structured tensor provides position guidance for the downstream policy.

\subsection{Locally Constrained Autoregressive Action Policy}

Within the MDP formulation, the action space $\mathcal A$ determines how the UAV changes the current reconstruction state. To simplify 3D viewpoint search and maintain motion continuity between adjacent views, we define a locally constrained discrete action $a_t=(a_t^{\mathrm{pos}},a_t^{\mathrm{ori}},b_t^{\mathrm{stop}})$. It comprises the position action $a_t^{\mathrm{pos}}=(\Delta x_t,\Delta y_t,\Delta z_t)$, the orientation action $a_t^{\mathrm{ori}}=(\psi_t,\phi_t)$, and the binary termination decision $b_t^{\mathrm{stop}}$. Here, $\Delta x_t$, $\Delta y_t$, and $\Delta z_t$ denote displacements relative to the current UAV position, whereas $\psi_t$ and $\phi_t$ denote the target yaw and pitch angles, respectively. The stop action allows the agent to terminate scanning once it determines that the reconstruction task is sufficiently complete.

Constraining motion to local relative displacements reduces the single-step position-search space and simplifies policy optimization. Meanwhile, relative actions allow the UAV to progressively explore a workspace beyond any fixed global candidate set. This design also discourages large jumps and repeated backtracking, produces smoother transitions between successive viewpoints, and facilitates fine-grained scanning of occluded regions, blind spots, and geometrically complex structures.

As shown in Fig.~\ref{fig:framework}, the position and termination decisions are jointly predicted from the shared representation $h_t$, i.e., $a_t^{\mathrm{pos}},b_t^{\mathrm{stop}}\sim\pi_{\mathrm{pos-stop}}(a_t^{\mathrm{pos}},b_t^{\mathrm{stop}}\mid h_t)$. After the position action is determined, the target position is encoded into a position embedding $e_t^{\mathrm{pos}}$. The orientation branch then predicts the camera direction conditioned on the shared state and the selected position as $a_t^{\mathrm{ori}}\sim\pi_{\mathrm{ori}}(a_t^{\mathrm{ori}}\mid h_t,e_t^{\mathrm{pos}})$.
Together, the position--termination and orientation branches define the MDP policy $\pi(a_t\mid s_t)$. The policy therefore captures a conditional sequence: first deciding where to move and then where to look upon arrival.

\subsection{Reward Design and Policy Optimization}
\label{sec:reward}

Within the MDP, the reward function $\mathcal R(s_t,a_t,s_{t+1})$ evaluates how the selected action changes the reconstruction state. We train the policy using proximal policy optimization (PPO)~\cite{schulman2017ppo} in the simulation environment and design the reward based on the principle that geometrically complex regions should be observed from a more diverse set of directions. Specifically, we use directional coverage gain as the primary reward, together with step, path-length, and terminal terms.

Complexity-weighted directional coverage is defined as
\begin{equation}
    C_t^{\mathrm{dir}}
    =
    \frac{
        \sum_v
        c_v
        \sum_j
        M(v,j)B_t(v,j)
    }{
        \sum_v
        c_v
        \sum_j
        M(v,j)
    },
    \label{eq:directional-coverage}
\end{equation}
where $B_t(v,j)$ indicates whether direction $j$ of voxel $v$ has been covered, $M(v,j)$ masks unobservable directions, and $c_v$ denotes the geometric complexity. Weighting coverage by $c_v$ encourages more diverse observations of complex regions. Both $M(v,j)$ and $c_v$ are computed from ground-truth data and used exclusively for reward computation rather than as policy inputs.

As coverage increases, obtaining new valid observation directions becomes progressively more difficult. To strengthen the learning signal during the later stages of scanning, we apply a monotonically increasing convex function to nonlinearly shape the directional-coverage increment:
\begin{equation}
    r_t^{\mathrm{dir}}
    =
    f\left(C_{t+1}^{\mathrm{dir}}\right)
    -
    f\left(C_{t}^{\mathrm{dir}}\right),
    \label{eq:nonlinear-reward}
\end{equation}
where $f(\cdot)$ is a convex function that increases monotonically. This design assigns a larger reward to the same coverage increment at higher coverage levels, encouraging the policy to continue searching for directions that are difficult to observe.

The complete reward function is
\begin{equation}
    r_t=
    \lambda_{\mathrm{dir}}r_t^{\mathrm{dir}}
    -
    \lambda_{\mathrm{len}}\Delta l_t
    -
    \lambda_{\mathrm{step}}
    +
    r_t^{\mathrm{term}},
    \label{eq:total-reward}
\end{equation}
where $\Delta l_t$ is the path-length increment, and $\lambda_{\mathrm{step}}$ is the per-step sensing and decision cost. A proper termination decision is rewarded according to the final reconstruction quality, whereas the policy is penalized for collisions, or producing excessively long trajectories.

The resulting policy is optimized to maximize the expected discounted return with discount factor $\gamma$. We train DA-NBV using PPO in multiple parallel simulation environments~\cite{schulman2017ppo}. Our implementation uses Stable-Baselines3 and follows the massively parallel training structure of Legged Gym~\cite{raffin2021stablebaselines3,rudin2022learning}. The value network estimates the current state value $V(s_t)$ from the shared representation $h_t$. During training, the policy collects trajectories and computes advantage estimates from the observed rewards and predicted state values. PPO then updates the policy using a clipped surrogate objective, which limits excessive deviations from the previous policy and improves training stability. The overall objective combines the policy loss, value prediction loss, and an entropy term that encourages exploration.

\subsection{Maritime Simulation Environment and Dataset}
\label{sec:environment}
For training and evaluation, we build a sea-state-aware ship-scanning simulation environment in NVIDIA Isaac Gym~\cite{makoviychuk2021isaacgym}. The simulated UAV is modeled after the Crazyflie 2.0 quadrotor platform~\cite{giernacki2017crazyflie}. For each sea-state level, we superimpose multiple wave components with physically sampled wavelengths, heights, and directions to generate the ocean surface, following spectrum-based and real-time irregular-wave modeling principles~\cite{pierson1964spectral,zheleznyakova2020shipmotion}. To efficiently approximate the wave-induced heave, roll, and pitch components of ship motion, we sample four points on the waterline plane: the longitudinal and lateral height differences determine pitch and roll, respectively, while their mean height determines heave. The time-varying ocean surface is also used to model occlusion of the lower hull caused by rising waves. In addition, we simulate a spatially varying wind field with sampled wind speed, direction, and vertical shear, and incorporate its influence on UAV motion and flight efficiency~\cite{liu2022wind}.

We further construct the SeaShip-3D dataset, which contains 300 ship models with diverse geometric structures and detailed textures. All models are calibrated according to their waterlines and normalized by ship length. Surface point clouds of the portions above the waterline are pre-sampled using Poisson-disk sampling implemented in Open3D for coverage and reconstruction-quality evaluation~\cite{yuksel2015poisson,zhou2018open3d}.

\section{Experiments}

\begin{table*}[t]
    \centering
    {
    \small
    \setlength{\tabcolsep}{1.5mm}
    \begin{tabular}{@{}l*{16}{c}@{}}
        \toprule
        Method
        & \multicolumn{4}{c}{SeaShip-3D (dynamic)}
        & \multicolumn{4}{c}{Houses3K (static)}
        & \multicolumn{4}{c}{OmniObject3D (static)}
        & \multicolumn{4}{c}{All} \\
        \cmidrule(lr){2-5}
        \cmidrule(lr){6-9}
        \cmidrule(lr){10-13}
        \cmidrule(lr){14-17}

        & CR$\uparrow$
        & CD$\downarrow$
        & $A_s\uparrow$
        & $A_p\uparrow$

        & CR$\uparrow$
        & CD$\downarrow$
        & $A_s\uparrow$
        & $A_p\uparrow$

        & CR$\uparrow$
        & CD$\downarrow$
        & $A_s\uparrow$
        & $A_p\uparrow$

        & CR$\uparrow$
        & CD$\downarrow$
        & $A_s\uparrow$
        & $A_p\uparrow$ \\
        \midrule

        Random
        & 55.15 & 22.08 & 51.22 & 50.50
        & 61.13 & 23.17 & 53.28 & 54.94
        & 54.59 & 32.13 & 49.67 & 48.41
        & 56.96 & 25.79 & 51.39 & 51.28 \\

        A-RMAP
        & 79.68 & 14.09 & 75.53 & 75.42
        & 78.29 & 15.74 & 75.03 & 74.70
        & 79.72 & 17.85 & 74.77 & 73.44
        & 79.23 & 15.89 & 75.11 & 74.52 \\

        Scan-RL
        & 83.07 & 14.11 & 76.59 & 77.21
        & 81.08 & 14.86 & 75.16 & 75.30
        & 82.85 & 17.34 & 75.09 & 76.94
        & 82.33 & 15.44 & 75.61 & 76.48 \\

        GenNBV
        & 90.20 & 9.97 & 86.62 & 85.26
        & 90.69 & 11.65 & 85.66 & 82.79
        & 89.09 & 13.97 & 83.56 & 83.76
        & 89.99 & 11.86 & 85.28 & 83.94 \\

        Hestia
        & 95.53 & 6.49 & 92.15 & 91.50
        & 96.03 & 6.18 & \textbf{94.04} & 91.37
        & 95.46 & 6.47 & \textbf{92.91} & 90.50
        & 95.67 & 6.38 & \textbf{93.03} & 91.12 \\

        \midrule

        DA-NBV
        & \textbf{98.49}
        & \textbf{3.68}
        & \textbf{93.04}
        & \textbf{94.35}

        & \textbf{98.28}
        & \textbf{3.29}
        & 93.02
        & \textbf{94.84}

        & \textbf{97.11}
        & \textbf{4.25}
        & 92.62
        & \textbf{94.37}

        & \textbf{97.96}
        & \textbf{3.74}
        & 92.89
        & \textbf{94.52} \\
        \bottomrule
    \end{tabular}
    }
    \caption{
        Quantitative comparison on SeaShip-3D, Houses3K, and
        OmniObject3D. The ``All'' columns report the arithmetic mean over
        the three datasets. Metrics are coverage rate (CR),
        Chamfer distance (CD), coverage--step area under the curve
        ($A_s$), and coverage--path-length area under the curve
        ($A_p$).
    }
    \label{tab:main-results}
\end{table*}

\begin{table}[t]
    \centering
    {
    \small
    \begin{tabular*}{\columnwidth}
        {@{\extracolsep{\fill}}ccc@{\hspace{2mm}}ccc@{}}
        \toprule
        \multicolumn{3}{c}{Components}
        & \multicolumn{3}{c}{Metrics} \\
        \cmidrule(lr){1-3}
        \cmidrule(lr){4-6}
        Reward
        & Action
        & PAF
        & CR$\uparrow$
        & DCR$\uparrow$
        & Dist.$\downarrow$ \\
        \midrule

        $\checkmark$
        &
        &
        &
        92.70 & 61.38 & 306.77 \\

        &
        $\checkmark$
        &
        &
        94.13 & 63.77 & 135.71 \\

        &
        &
        $\checkmark$
        &
        95.40 & 85.99 & 415.62 \\

        \midrule

        $\checkmark$
        & $\checkmark$
        &
        &
        95.51 & 68.12 & \textbf{131.76} \\

        $\checkmark$
        &
        & $\checkmark$
        &
        96.36 & 87.26 & 412.01 \\

        &
        $\checkmark$
        & $\checkmark$
        &
        97.40 & 90.47 & 207.29 \\

        \midrule

        $\checkmark$
        & $\checkmark$
        & $\checkmark$
        & \textbf{98.49}
        & \textbf{92.06}
        & 196.22 \\
        \bottomrule
    \end{tabular*}
    }
    \caption{
        Ablation study of the nonlinear reward design, the locally constrained autoregressive action policy, and the PAF. Metrics are coverage rate (CR), directional coverage rate (DCR), and travel distance (Dist.).
    }
    \label{tab:ablation}
\end{table}

We evaluate DA-NBV in both dynamic maritime and static
object-reconstruction environments. The main experiments are conducted
on SeaShip-3D under different sea states, with additional ablation and
cross-dataset experiments on Houses3K and OmniObject3D.

\subsection{Experimental Setup}
\label{sec:experimental-setup}

SeaShip-3D is randomly divided into mutually exclusive training and test sets for training and evaluation. Static-scene experiments are additionally conducted in the simulation environment using splits of Houses3K~\cite{peralta2020scanrl} and a 200-object subset of OmniObject3D~\cite{wu2023omniobject3d}, covering a wide variety of object categories and geometric structures. All objects are normalized to a maximum dimension of \(15\,\mathrm{m}\).

At each reset, we randomly sample the initial UAV pose, position,  
ship heading, and wind--wave conditions according to the selected
sea-state level. For fairness, all methods use the same coordinate-registration
procedure to compensate for ship motion. We evaluate all methods using the same fixed number of views.

The simulated camera has a resolution of $400 \times 400$ pixels and a vertical field of view of $90^{\circ}$. All experiments are conducted on an NVIDIA RTX 4090 GPU. All methods are trained and evaluated under identical conditions.

The main comparison in Table~\ref{tab:main-results} reports CR, CD, $A_s$, and $A_p$, whereas the ablation study in Table~\ref{tab:ablation} reports CR, DCR, and Dist. Coverage rate (CR, \%) is the fraction of ground-truth points within a distance threshold of the reconstruction~\cite{knapitsch2017tanks}. Chamfer distance (CD, cm) measures point-set discrepancy~\cite{fan2017pointset}. We compute CD and perform batched point-cloud processing using PyTorch3D~\cite{ravi2020pytorch3d}. Coverage--step area under the curve ($A_s$, \%)~\cite{chen2024gennbv} and coverage--path-length area under the curve ($A_p$, \%) quantify coverage efficiency with respect to the number of scanning steps and UAV travel distance, respectively. Directional coverage rate (DCR, \%) measures observation completeness across viewing directions, and UAV travel distance (Dist., m) measures the cumulative straight-line distance between consecutive selected viewpoints.

\subsection{Performance Comparison}
We compare DA-NBV with Random, ActiveRMAP (A-RMAP)~\cite{zhan2022activermap}, Scan-RL~\cite{peralta2020scanrl}, GenNBV~\cite{chen2024gennbv}, and Hestia~\cite{lu2026hestia}.
\paragraph{Quantitative comparison.}
As shown in Table~\ref{tab:main-results}, DA-NBV outperforms the baselines in reconstruction completeness, accuracy, and acquisition efficiency across the evaluated dynamic sea environment and both static datasets. Compared with existing methods, DA-NBV achieves greater surface coverage with higher path efficiency. Moreover, DA-NBV remains competitive on Houses3K and OmniObject3D, indicating its generalizability to different types of static objects.

\begin{figure*}
    \centering
    \includegraphics[width=\textwidth]{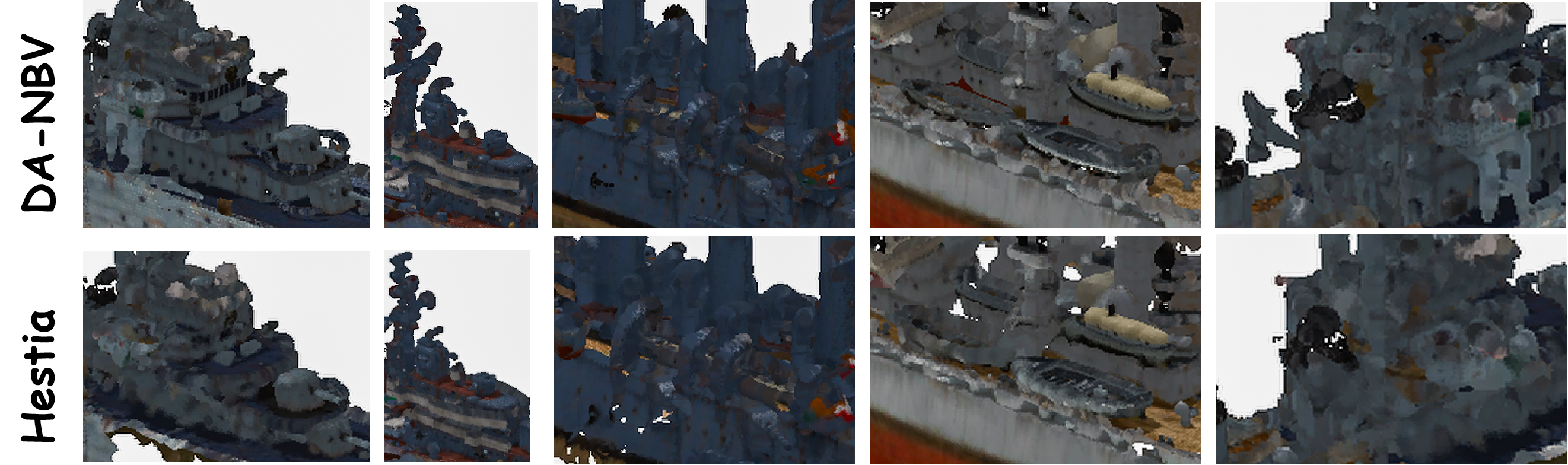}
    \caption{Qualitative comparison of reconstruction results. Both Hestia and DA-NBV achieve nearly complete reconstruction of the ship. However, DA-NBV preserves substantially finer geometric details with fewer missing regions. }
    \label{fig:reconstruction}
\end{figure*}

\paragraph{Qualitative comparison.}
We perform Screened Poisson surface reconstruction using Open3D~\cite{kazhdan2013screened,zhou2018open3d}. As shown in Fig.~\ref{fig:reconstruction}, DA-NBV produces more complete reconstructions with finer geometric details. Compared with other methods, it more reliably recovers complex structures, supporting the effectiveness of our strategy and the principle that geometrically complex regions should be observed from a more diverse set of directions.

\subsection{Ablation Studies}
\label{sec:ablation}

We conduct ablation studies to evaluate the contributions of the nonlinear reward design, the locally constrained autoregressive action policy, and the PAF. Across matched configurations in Table~\ref{tab:ablation}, the PAF consistently increases DCR by guiding the policy toward regions with insufficient directional observations. The action policy consistently reduces Dist. by discouraging large viewpoint transitions and repeated backtracking. The nonlinear reward improves CR, DCR, and Dist. by strengthening the learning signal for difficult regions and missing directions.

These trends reveal complementary roles: the PAF primarily improves directional coverage, the action policy enhances path efficiency, and the nonlinear reward helps the policy exploit both designs. The complete model achieves the highest CR and DCR, whereas the reward--action variant yields the shortest travel distance.

\begin{figure}[t]
    \centering
    \includegraphics[width=\columnwidth]{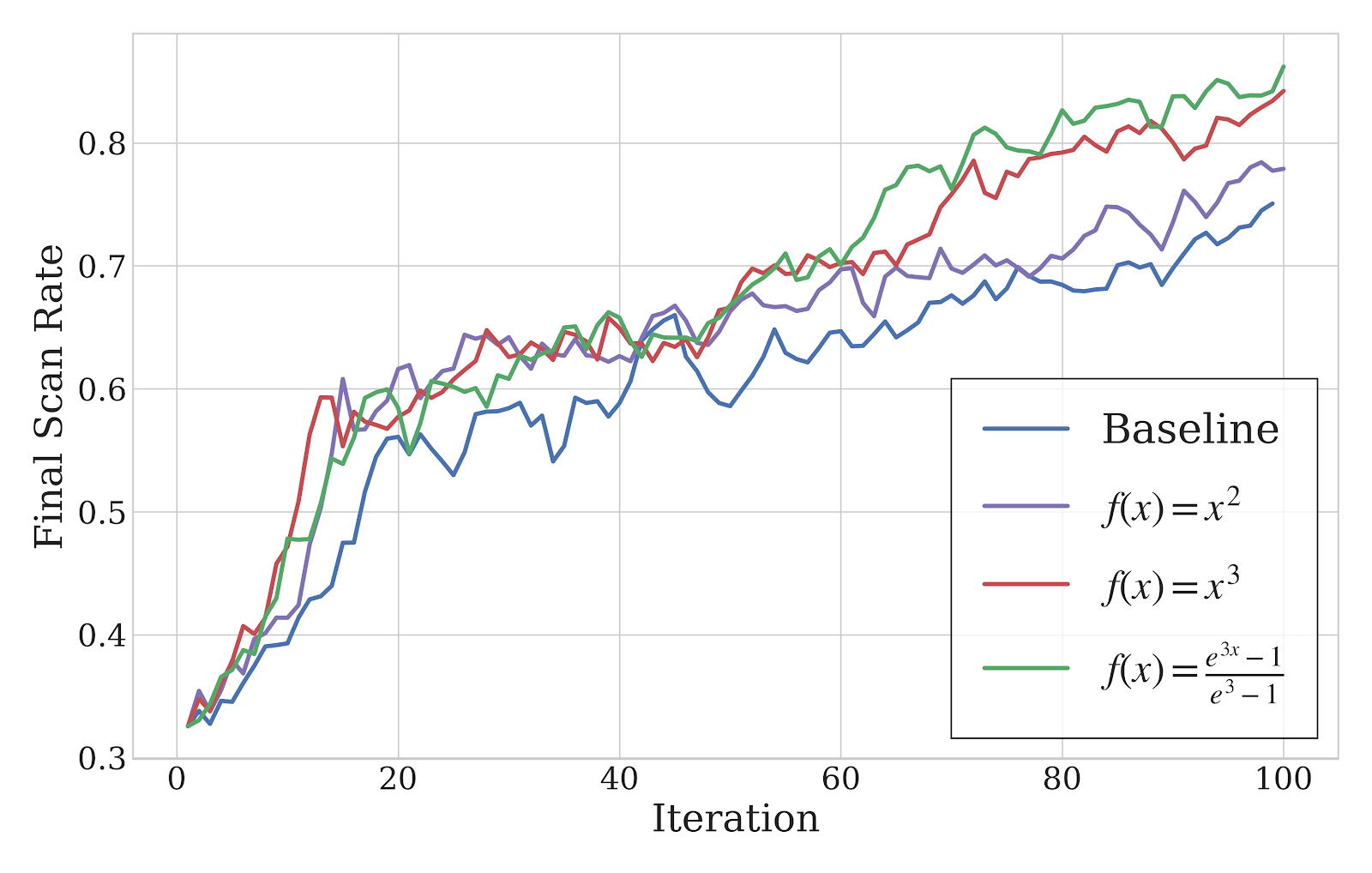}
    \caption{Comparison of different shaping functions for the directional-coverage reward.}
    \label{fig:reward}
\end{figure}

\subsection{Additional Analysis}

\begin{table}[t]
    \centering
    {
    \small
    \setlength{\tabcolsep}{0.6mm}
    \renewcommand{\arraystretch}{1.0}
    \begin{tabular}{@{}lcccccccc@{}}
        \toprule
        Method
        & \multicolumn{4}{c}{Sea State 0}
        & \multicolumn{4}{c}{Sea State 9} \\
        \cmidrule(lr){2-5}
        \cmidrule(lr){6-9}

        & CR$\uparrow$
        & CD$\downarrow$
        & $A_s\uparrow$
        & $A_p\uparrow$
        & CR$\uparrow$
        & CD$\downarrow$
        & $A_s\uparrow$
        & $A_p\uparrow$ \\
        \midrule

        Random
        & 56.57 & 20.93 & 51.92 & 52.62
        & 56.91 & 22.82 & 52.23 & 49.65 \\

        A-RMAP
        & 80.61 & 14.28 & 75.54 & 75.04
        & 76.60 & 15.71 & 73.17 & 74.50 \\

        Scan-RL
        & 83.75 & 14.01 & 77.22 & 78.30
        & 80.52 & 14.99 & 75.10 & 76.71 \\

        GenNBV
        & 91.15 & 8.86 & 87.08 & 85.81
        & 88.14 & 11.27 & 86.93 & 85.19 \\

        Hestia
        & 96.51 & 6.03 & 92.76 & 90.93
        & 94.14 & 7.16 & 91.37 & 90.67 \\

        \midrule

        DA-NBV
        & \textbf{98.75}
        & \textbf{3.44}
        & \textbf{93.57}
        & \textbf{95.19}
        & \textbf{97.86}
        & \textbf{4.32}
        & \textbf{91.63}
        & \textbf{93.66} \\
        \bottomrule
    \end{tabular}
    }
    \caption{
        Quantitative comparison on SeaShip-3D under calm
        (Sea State 0) and severe (Sea State 9) maritime conditions.
        Metrics are coverage rate (CR), Chamfer distance (CD),
        coverage--step area under the curve ($A_s$), and
        coverage--path-length area under the curve ($A_p$).
    }
    \label{tab:sea-state-results}
\end{table}

\paragraph{Policy optimization efficiency.}
We further analyze the effects of nonlinear reward shaping. As shown in Fig.~\ref{fig:reward}, multiple convex shaping functions consistently outperform the linear formulation with steeper learning curves, exhibiting better final performance and faster convergence. These results validate the effectiveness of our reward-shaping design.

\paragraph{Computational efficiency.}
To assess computational efficiency, we benchmark the complete DA-NBV decision-making pipeline, excluding the Isaac Gym simulation. Across three runs of 1,000 iterations, DA-NBV achieves an average decision latency of \(8.697 \pm 0.324\) ms, demonstrating real-time computational feasibility on the evaluated hardware.

\paragraph{Robustness to dynamic maritime conditions.}
As shown in Table~\ref{tab:sea-state-results}, a comparison of Sea States 0 and 9 shows that DA-NBV exhibits limited performance degradation under severe maritime disturbances. Its CR decreases by only \(0.89\) percentage points, while CD increases by only \(0.88\) cm. These changes are smaller than those of the strongest reconstruction baselines, Hestia and GenNBV. DA-NBV also maintains high efficiency under Sea State 9, achieving \(91.63\%\) \(A_s\) and \(93.66\%\) \(A_p\). These results demonstrate that DA-NBV remains effective under maritime disturbances.

\section{Conclusion}
We present DA-NBV, a direction-aware next-best-view planner for efficient 3D reconstruction of ships. By incorporating voxel-level directional observations, a Position Advantage Field, a reward design aligned with practical scanning objectives, and a locally constrained autoregressive policy, DA-NBV integrates directional awareness into state representation, viewpoint scoring, and action selection. Experiments on SeaShip-3D and the static datasets show that DA-NBV outperforms the baselines in reconstruction completeness and accuracy while achieving higher path efficiency. Future work will focus on real-world UAV deployment.

\bibliography{reference}

\clearpage

\begin{center}
    {\LARGE\bf Supplementary Material}
\end{center}

\vspace{1em}

\section{Implementation Details}

\subsection{Network Inputs and Action Heads}
The policy uses three complementary state streams. The occupancy branch encodes the single-channel 3D occupancy grid with two 3D convolutional layers into a 256-dimensional geometric embedding. The motion branch processes the recent relative-pose and action history using sinusoidal positional encoding followed by two MLP layers; the current relative pose is encoded separately as a 64-dimensional linear embedding and fused with the history feature. The Position Advantage Field (PAF) branch encodes a single-channel scalar field $\mathbf U_t$ defined over the local 3D candidate-position lattice. Two 3D convolutional layers map $\mathbf U_t$ to a 128-dimensional feature. The three embeddings are concatenated and mapped to a 256-dimensional shared representation used by both the PPO policy and value networks.

The learnable position-advantage scorer (LPAS) constructs this field from voxel--candidate pairs. For a candidate position $q$ and an active voxel $v$, its input combines voxel-state attributes---position, observation support and count, geometric complexity, and the coverage states of 12 directional bins---with candidate-dependent distance, visibility, and missing-direction alignment. Its scalar pairwise output is
\begin{equation}
\ell(v,q)=\mathbf w_2^\top\sigma\!\left(\mathbf W_1\mathbf g(v,q)+\mathbf b_1\right)+b_2,
\label{eq:supp_lpas}
\end{equation}
where $\mathbf g(v,q)$ denotes the concatenated voxel--candidate features and $\sigma$ is a nonlinear activation. Aggregating $\ell(v,q)$ over the active, visibility-valid voxel set $\mathcal V_t^{\mathrm{valid}}$ gives one scalar advantage for each candidate,
\begin{equation}
U_t(q)=\sum_{v\in\mathcal V_t^{\mathrm{valid}}}\ell(v,q).
\label{eq:supp_paf_aggregation}
\end{equation}
Placing the values $U_t(q)$ at their corresponding coordinates produces the single-channel PAF $\mathbf U_t\in\mathbb R^{N_x\times N_y\times N_z}$. Thus, the voxel attributes are LPAS inputs, whereas the PAF contains one aggregated scalar per candidate position and guides translation selection in the first action stage.

Importantly, the policy does not greedily select the candidate with the largest instantaneous advantage $U_t(q)$. Such a rule maximizes one-step expected coverage gain and can therefore improve the coverage--step area under the curve ($A_s$). However, a locally high-value candidate can be far from the current viewpoint or on the opposite side of the target. Repeatedly pursuing these isolated maxima can produce long, back-and-forth trajectories and overlook nearby viewpoints that jointly expose the remaining surface. Consequently, strong per-step gain need not yield strong coverage per unit travel, and the coverage--path-length area under the curve ($A_p$) can be low. Instead, the policy concatenates the encoded single-channel PAF with occupancy-grid and motion-history features before action prediction. This joint representation enables it to trade off immediate geometric gain, travel cost, spatial continuity, and the reachability of useful future viewpoints, thereby making long-horizon decisions rather than repeatedly selecting the largest one-step advantage.

The policy factorizes the discrete action into two stages. The first stage jointly predicts the relative translation $(x_t,y_t,z_t)$ and the binary stop decision. Conditioned on the selected translation, the attitude head predicts pitch and yaw; roll is fixed to zero. Let $\boldsymbol{\rho}_t=(x_t,y_t,z_t)$ and define the element-wise transformed translation $\tilde{\boldsymbol{\rho}}_t=\operatorname{asinh}(\boldsymbol{\rho}_t)$. We encode it with two Fourier frequencies $\omega_1$ and $\omega_2$ and concatenate the result with the state feature $s_t$:
\begin{equation}
\mathbf z_t=\left[s_t;\tilde{\boldsymbol{\rho}}_t;
\sin(\omega_1\tilde{\boldsymbol{\rho}}_t);\cos(\omega_1\tilde{\boldsymbol{\rho}}_t);
\sin(\omega_2\tilde{\boldsymbol{\rho}}_t);\cos(\omega_2\tilde{\boldsymbol{\rho}}_t)\right].
\label{eq:supp_attitude_encoding}
\end{equation}
The attitude distribution is therefore parameterized as $\pi(\mathrm{pitch}_t,\mathrm{yaw}_t\mid\mathbf z_t)$. The complete factorization is
\begin{equation}
\begin{aligned}
\pi(a_t\mid o_t)={}&\pi(x_t,y_t,z_t,\mathrm{stop}_t\mid s_t)\\
&\pi(\mathrm{pitch}_t,\mathrm{yaw}_t\mid \mathbf z_t),
\end{aligned}
\end{equation}
with $a_t=[x_t,y_t,z_t,0,\mathrm{pitch}_t,\mathrm{yaw}_t,\mathrm{stop}_t]$.

\subsection{Visibility Approximation}
Exact ray tracing over every voxel--candidate pair is too costly during planning. We therefore estimate visibility from the axis-aligned bounding box of the accumulated point cloud in each occupied voxel, as described in the main paper. When aggregating the candidate advantage in Eq.~\eqref{eq:supp_paf_aggregation}, a voxel contribution is masked when its estimated visibility is below $0.10$. The continuous visibility value is nevertheless retained as an LPAS input, allowing the scorer to distinguish partially visible candidates from clearly visible ones.

For an occupied voxel $u$, let $\Delta_k(u)=x_{u,k}^{\max}-x_{u,k}^{\min}$ denote its extent along axis $k$. Let $A_{ij}$ be the area of the voxel face orthogonal to axis $k$, where $\{i,j,k\}=\{x,y,z\}$. The axis-aligned transmittance is
\begin{equation}
\tau_k(u)=1-\frac{\Delta_i(u)\Delta_j(u)}{A_{ij}}.
\label{eq:supp_axis_transmittance}
\end{equation}
For an arbitrary unit viewing direction $\mathbf d$, we interpolate these estimates using the magnitudes of the directional components:
\begin{equation}
\tau(u,\mathbf d)=
\frac{\sum_{k\in\{x,y,z\}}|d_k|\tau_k(u)}
{\sum_{k\in\{x,y,z\}}|d_k|+\varepsilon}.
\label{eq:supp_directional_transmittance}
\end{equation}
Given a target voxel $v$ and a candidate position $q$, we sample occupied voxels $\mathcal R(v,q)$ along the line segment from $v$ to $q$ and estimate the path visibility as
\begin{equation}
\operatorname{Vis}(v,q)=
\prod_{u\in\mathcal R(v,q)}\tau\bigl(u,d(v,q)\bigr).
\label{eq:supp_path_visibility}
\end{equation}

We validated the visibility approximation on 200 objects drawn from the ship, house, and everyday-object datasets, using 20 viewpoints uniformly sampled from the object-centered hemisphere for each object. Visibility labels were obtained by back-projecting rendered depth images. At the $0.10$ threshold, 88.3\% of truly invisible voxels were masked, whereas 13.4\% of truly visible voxels were falsely masked. These errors were concentrated in small, marginally visible regions and had limited influence on action selection.

\subsection{Geometric Complexity and Reward Shaping}

\subsubsection{Shared PCA Descriptor Definitions}
For an active voxel $v$, let $\mathcal{P}_t(v)=\{p_i\}_{i=1}^{m_t(v)}$ denote its accumulated local point set at time $t$, where $m_t(v)$ is the number of fused 3D points. We compute the centroid and covariance matrix as
\begin{equation}
\begin{aligned}
\bar{p}_t(v)&=\frac{1}{m_t(v)}\sum_{i=1}^{m_t(v)}p_i,\\
\mathbf{\Sigma}_t(v)&=\frac{1}{m_t(v)}\sum_{i=1}^{m_t(v)}
\Bigl(p_i-\bar{p}_t(v)\Bigr)\Bigl(p_i-\bar{p}_t(v)\Bigr)^\top.
\end{aligned}
\label{eq:supp_covariance}
\end{equation}
Let $\lambda_1\geq\lambda_2\geq\lambda_3\geq0$ be the eigenvalues of $\mathbf{\Sigma}_t(v)$. We define the PCA descriptors as
\begin{equation}
\begin{aligned}
L_t(v)&=\frac{\lambda_1-\lambda_2}{\lambda_1+\varepsilon},\\
S_t(v)&=\frac{\lambda_3}{\lambda_1+\varepsilon},\\
C_t(v)&=\frac{\lambda_3}{\lambda_1+\lambda_2+\lambda_3+\varepsilon},
\end{aligned}
\label{eq:supp_pca_features}
\end{equation}
where $L_t$, $S_t$, and $C_t$ are linearity, scattering, and curvature, respectively, and $\varepsilon$ is a small constant for numerical stability. Linearity is large for locally one-dimensional structures; scattering increases when the point distribution spreads across all three principal directions; curvature measures the relative variation along the local normal direction. The following two branches use this shared descriptor definition but never use the same point set: online descriptors are computed from accumulated back-projected observations as voxel-side LPAS inputs, whereas ground-truth descriptors are computed from ground-truth surface points only for reward weighting and visualization.

\subsubsection{Ground-truth Reward Complexity}
The following scalar is computed from the ground-truth descriptors, rather than from online observations, and is used only for reward construction and evaluation-side visualization:
\begin{equation}
c^{\mathrm{GT,raw}}(v)=S^{\mathrm{GT}}(v)+C^{\mathrm{GT}}(v)-L^{\mathrm{GT}}(v).
\label{eq:supp_raw_complexity}
\end{equation}
For reward construction, the ground-truth score is piecewise-linearly remapped over all voxels of a ship to the interval $[1,3]$, yielding the directional-coverage weight $c_v$. This strictly positive weighting preserves a contribution from simple regions while assigning up to three times the weight to geometrically complex ones.

\begin{figure*}[p]
\centering
\includegraphics[width=\textwidth,height=0.86\textheight,keepaspectratio]{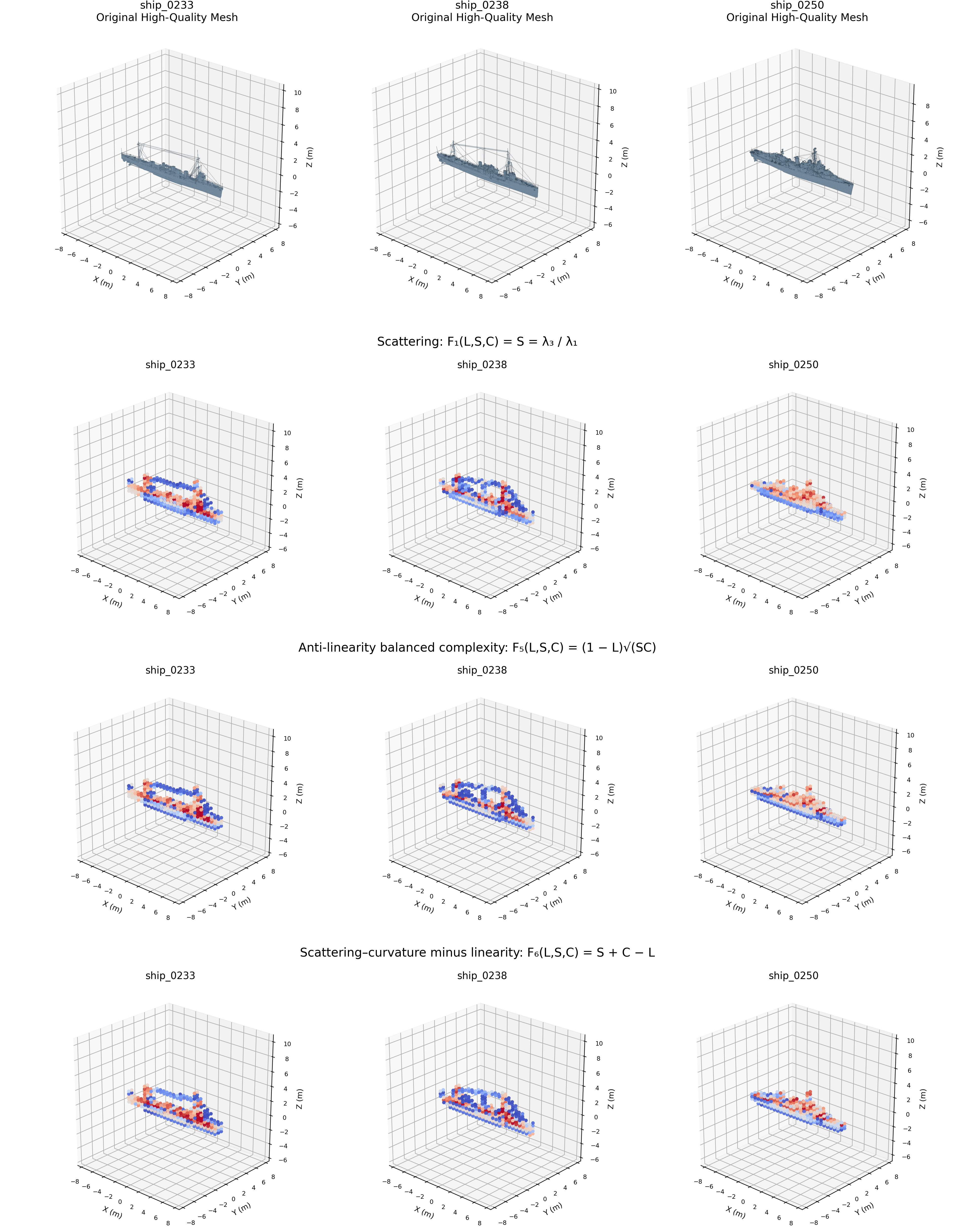}
\caption{Ground-truth complexity visualizations used for reward-side analysis. From top to bottom: representative high-quality meshes, scattering maps, and combined complexity maps derived from scattering, curvature, and linearity.}
\label{fig:supp_complexity}
\end{figure*}

\begin{figure*}[p]
\centering
\includegraphics[height=0.84\textheight,keepaspectratio]{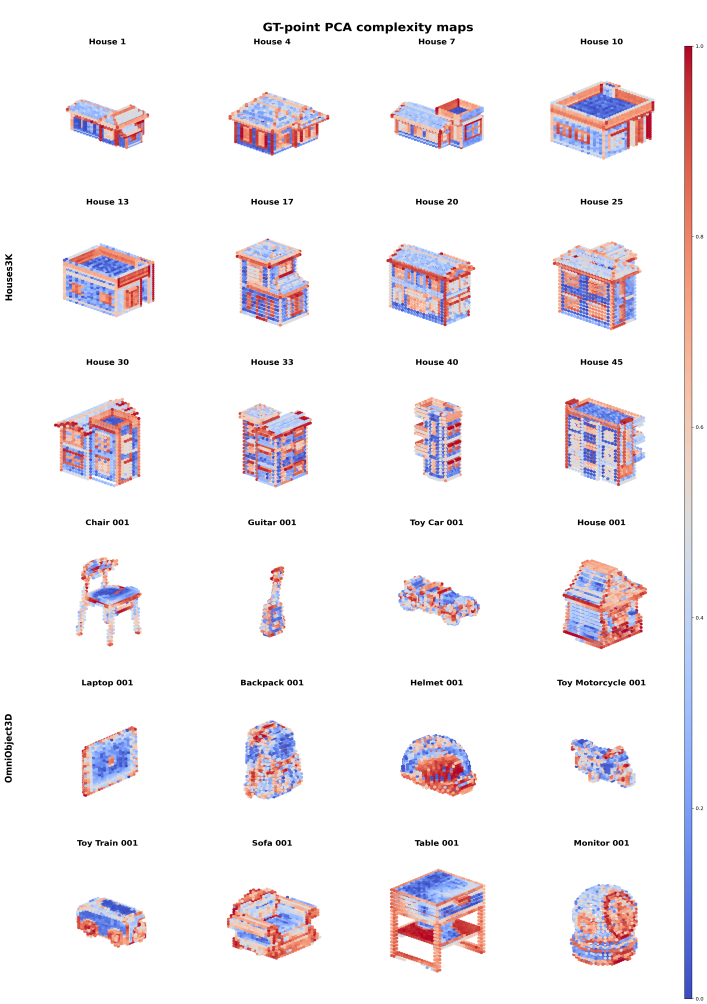}
\caption{PCA-derived complexity maps computed from ground-truth surface points for the cross-category generalization data. The upper three rows show Houses3K examples; the lower three rows show OmniObject3D examples. Colors indicate the normalized complexity score.}
\label{fig:supp_cross_category_complexity}
\end{figure*}

\subsubsection{Online Complexity for LPAS Inputs}
For the online branch, the preceding PCA equations are applied to the accumulated back-projected point set, yielding $L_t^{\mathrm{raw}}(v)$, $S_t^{\mathrm{raw}}(v)$, and $C_t^{\mathrm{raw}}(v)$. Online estimates from very few points are unreliable: a small number of outliers can produce misleading PCA descriptors. We therefore attenuate each online descriptor by a monotone observation-confidence factor,
\begin{equation}
\alpha\bigl(N_t^{\mathrm{obs}}(v)\bigr)=1-\exp\!\left(-\frac{N_t^{\mathrm{obs}}(v)}{\kappa}\right),
\qquad \kappa=1,
\label{eq:supp_confidence_complexity}
\end{equation}
\begin{equation}
\hat L_t(v)=\alpha\bigl(N_t^{\mathrm{obs}}(v)\bigr)L_t^{\mathrm{raw}}(v),
\end{equation}
\begin{equation}
\hat S_t(v)=\alpha\bigl(N_t^{\mathrm{obs}}(v)\bigr)S_t^{\mathrm{raw}}(v),
\end{equation}
\begin{equation}
\hat C_t(v)=\alpha\bigl(N_t^{\mathrm{obs}}(v)\bigr)C_t^{\mathrm{raw}}(v).
\end{equation}
Here, $N_t^{\mathrm{obs}}(v)$ is the number of depth observations that contribute at least one fused point to $v$, which is distinct from the fused-point count $m_t(v)$ used in the covariance estimate. With $\kappa=1$, the confidence values after one, two, and three observations are approximately $0.632$, $0.865$, and $0.950$, respectively. The corrected online descriptors are used exclusively as voxel-side LPAS inputs and affect the PAF only through the learned pairwise utilities; the ground-truth branch above is used exclusively for reward computation.

\subsubsection{Unobservable-direction Mask}
For each voxel--direction pair $(v,j)$, we precompute an observability mask $M(v,j)$ from the ground-truth geometry. We uniformly sample a ray set $\mathcal R_{v,j}$ within a virtual camera frustum centered on voxel $v$ and oriented along the center of bin $j$, and perform ray casting against the ground-truth mesh. Let $\operatorname{first}(r)$ denote the surface element at the first hit of ray $r$, and let $\mathcal S(v)$ be the ground-truth local surface associated with $v$. We set
\begin{equation}
M(v,j)=\mathbb I\!\left[\exists r\in\mathcal R_{v,j}:\operatorname{first}(r)\in\mathcal S(v)\right].
\label{eq:supp_visibility_mask}
\end{equation}
Thus, a direction is valid only if at least one ray first hits the target local surface; directions fully blocked by other ship parts or the sea surface are excluded. The complexity-weighted directional coverage is
\begin{equation}
C_t^{\mathrm{dir}}=
\frac{\sum_v c_v\sum_j M(v,j)B_t(v,j)}
{\sum_v c_v\sum_j M(v,j)}.
\end{equation}

\subsubsection{Nonlinear Reward Shaping}
In the main experiments, we apply the nonlinear shaping function $f(x)=x^3$ to the coverage increment,
\begin{equation}
r_t^{\mathrm{dir}}=f\bigl(C_{t+1}^{\mathrm{dir}}\bigr)-f\bigl(C_t^{\mathrm{dir}}\bigr).
\end{equation}
Because $f$ is convex and monotone, an equal coverage increment receives a larger reward at higher coverage. The design principle is that completing a more difficult part of the task should earn more reward. Coverage difficulty grows nonlinearly: late in an episode, most readily accessible viewpoints have already been used, and finding a new valid viewing direction becomes substantially harder. The shaping therefore counters the tendency to terminate after easy regions are covered and encourages continued search for the remaining difficult directions.

\subsubsection{Voxel-level Direction Bins}
We discretize directional state into 12 spherical bins. Let $\mathbf b_j$ denote the unit center direction of bin $j$. Each valid viewing direction is assigned to the single bin with the greatest directional alignment:
\begin{equation}
\operatorname{bin}(\mathbf d)=
\arg\max_{j\in\{1,\ldots,12\}}\mathbf d^\top\mathbf b_j.
\end{equation}
Each bin therefore represents a finite angular region rather than a single exact vector. Views that remain within the same region after small roll or pitch perturbations activate the same bin, providing robustness to small vessel oscillations without removing the incentive for multi-directional coverage.
We use 12 approximately equal-area directional bins. Each bin has nominal solid angle $4\pi/12=\pi/3$ sr; the equal-area spherical-cap radius is $\theta_{\mathrm{eq}}=\arccos(5/6)\approx33.6^\circ$, equivalently a center-alignment threshold of $\cos\theta_{\mathrm{eq}}=5/6\approx0.833$. The exact Voronoi boundary is set by the fixed bin centers; this equivalent angle quantifies the angular resolution. Fewer bins do not adequately promote multi-directional scanning, whereas substantially more bins make the state unnecessarily sparse and redundant.

\section{Experimental Protocol}

\subsection{Baselines and Motion Compensation}
\textbf{Random} samples 5-DoF viewpoint actions uniformly from the same action domain and serves as a no-information reference. \textbf{ActiveRMAP} is a neural-radiance-field method for active mapping and planning that evaluates candidate views using the current neural scene representation. \textbf{Scan-RL} learns a next-best-view policy for active 3D reconstruction. For ActiveRMAP and Scan-RL, we sample actions on the ship-centered hemisphere and transform them using the vessel's instantaneous roll and pitch.

\textbf{GenNBV} is a generalizable policy for active 3D reconstruction based on a spatial reconstruction state. 
\textbf{Hestia} adopts a voxel-face-aware hierarchical acquisition strategy that explicitly models the observation states of individual voxel faces. 
For GenNBV, Hestia, and DA-NBV, we apply the same motion compensation procedure to their respective reconstruction state representations before viewpoint planning. 
This ensures that the performance differences are primarily attributed to the planning policies rather than motion-induced ghost occupancy.

\subsection{Coverage Metric and Threshold}
Let $P_{\mathrm{GT}}$ and $P_{\mathrm{rec}}$ denote the ground-truth and reconstructed surface point sets. We compute coverage rate as the ground-truth-to-reconstruction recall
\begin{equation}
\begin{aligned}
\mathrm{CR}&=\frac{1}{|P_{\mathrm{GT}}|}
\sum_{p\in P_{\mathrm{GT}}}
\mathbb{I}\!\left[\min_{q\in P_{\mathrm{rec}}}\lVert p-q\rVert_2<\tau\right],\\
\tau&=4\,\mathrm{cm}.
\end{aligned}
\label{eq:supp_cr}
\end{equation}
where $\mathbb{I}[\cdot]$ is the indicator function. We use $\tau=4$ cm to balance camera resolution against avoiding substantial geometric error. In an ideal efficient acquisition, the complete 15 m vessel is captured within the 400-pixel image dimension, giving the limiting sampling scale $15/400=3.75$ cm per pixel. We therefore choose a slightly larger 4 cm tolerance: it avoids an unrealistically subpixel criterion while limiting the accepted geometric deviation. The same threshold is used for every method and evaluation split.

\subsection{Evaluation Splits and Efficiency Metrics}
SeaShip-3D is randomly divided into mutually exclusive training and test sets. All learning-based methods are trained on the training split and evaluated on unseen test ships. The static-scene experiments use an object-disjoint 50-model test split of Houses3K and a 200-object subset of OmniObject3D.

\textbf{Dynamic maritime protocol.} For each sea-state level, we evaluate 50 test ships, with 10 independently randomized episodes per ship. Each reset samples an initial UAV pose, ship heading, and wind--wave conditions according to the selected sea-state level. The ship undergoes the wave-induced heave, roll, pitch, and translation described below, and all methods use the same coordinate-registration procedure to compensate for this motion before planning.

\textbf{Static-object protocol.} We evaluate 50 Houses3K models and 200 OmniObject3D objects, with 10 independently randomized episodes per object. Each episode begins from an independently sampled initial UAV pose on the upper hemisphere around the target. The same camera model, relative action domain, candidate-grid construction, and fixed view budget are used for all methods. Under each protocol, all methods are evaluated under matched conditions.

In addition to CR and Chamfer distance (CD), the main comparison reports the coverage--step area under the curve ($A_s$) and the coverage--path-length area under the curve ($A_p$). These metrics quantify coverage efficiency with respect to the number of scanning steps and UAV travel distance, respectively. The ablation study further reports directional coverage rate (DCR), which measures observation completeness across viewing directions, and travel distance (Dist.), defined as the cumulative straight-line distance between consecutive selected viewpoints.

\subsection{Training and Action Settings}
We train with 256 parallel environments for 1,000 iterations. PPO uses 128 rollout steps, 5 optimization epochs, a batch size of 128, a learning rate of $10^{-4}$, and seed 1. All networks are randomly initialized and trained end-to-end; training takes approximately 20 hours on an RTX 4090. The reference-scale relative translation action range is $\pm5$ m along each axis, with a 0.2 m step. The candidate lattice $\mathcal Q_t$ is a uniformly sampled $10\times10\times10$ grid with 1 m spacing, and the PAF is evaluated on this lattice. At runtime, these UAV-related lengths are multiplied by the ship-specific normalization coefficient defined under \emph{Dataset Construction and Scale Normalization}.

\begin{figure*}[!t]
\centering
\includegraphics[width=0.90\textwidth]{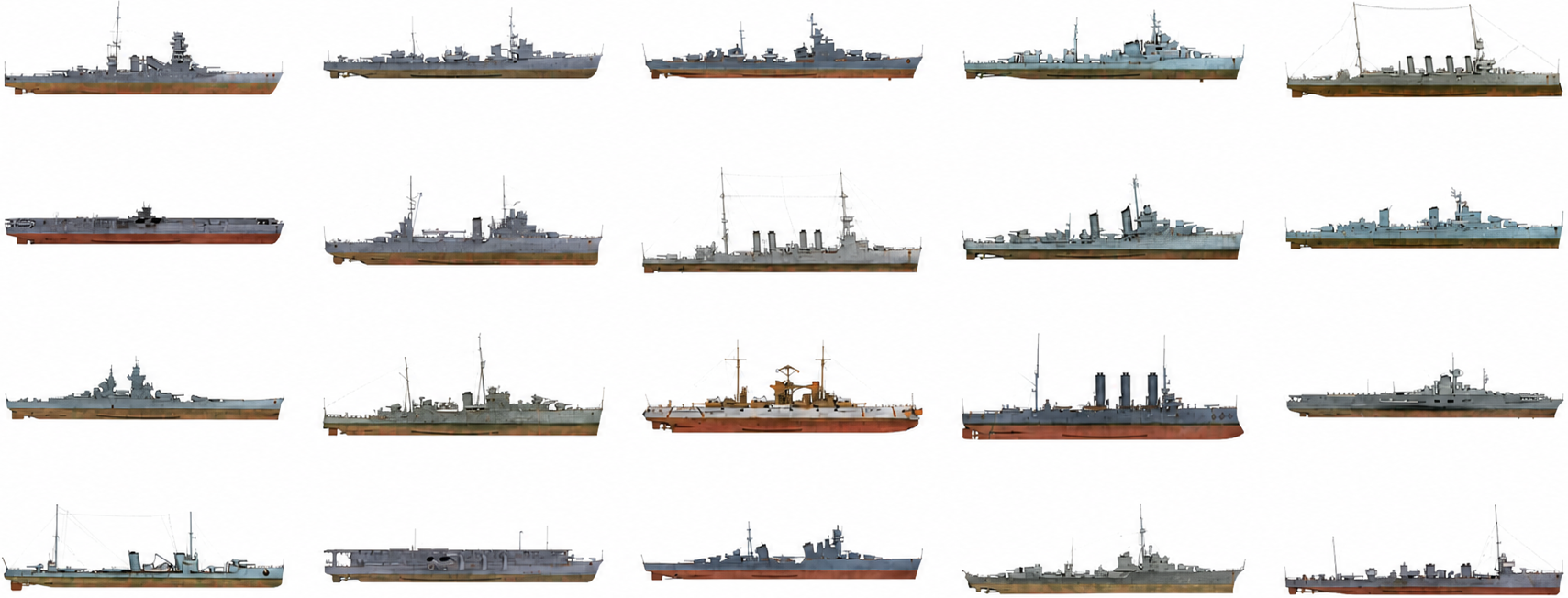}
\caption{Examples from the SeaShip-3D dataset after waterline alignment and scale normalization.}
\label{fig:supp_seaship}
\end{figure*}

\section{Maritime Simulation Details}

\subsection{Dataset Construction and Scale Normalization}
\textbf{SeaShip-3D.} The dataset contains 300 ship models. To place each model consistently with the water surface, we vertically position the still-water plane so that the portion below it contains 25\% of the ship mesh volume; this defines the waterline alignment. For a ship with original length $L_{\mathrm{orig}}$, we normalize its length to $L_{\mathrm{norm}}=15\,\mathrm{m}$ and retain the ship-specific normalization coefficient
\begin{equation}
\eta_L=\frac{L_{\mathrm{norm}}}{L_{\mathrm{orig}}}.
\label{eq:supp_length_scale}
\end{equation}
Every length-valued quantity in the maritime environment is transformed from its reference-scale value according to $\ell^{\mathrm{sim}}=\eta_L\ell^{\mathrm{ref}}$. The same coefficient is applied to the ship geometry, wave amplitudes and wavelengths, UAV standoff and action distances, and candidate-lattice spacing. Consequently, the relative geometry between the ship, wave field, and UAV trajectory is preserved after normalization. We uniformly sample $10^6$ points from the portion of each normalized mesh above the waterline using Poisson-disk sampling; these dense ground-truth points are used for coverage evaluation, visibility labels, and complexity supervision.

\textbf{Cross-category generalization.} To assess whether the method is specific to ship geometry, we additionally use 50 house models from Houses3K and 200 everyday objects from OmniObject3D. Each non-ship object is normalized to a longest axis of $15\,\mathrm{m}$, keeping the motion scale and reconstruction precision comparable across categories.

\subsection{Wave Field, Sea States, and Sea-surface Occlusion}
For each episode, we randomly sample a sea-state level $s\in\{0,\ldots,9\}$. The water surface is constructed from eight traveling sinusoidal components,
\begin{equation}
\begin{aligned}
h(x,y,t)&=\sum_{i=1}^{8} A_i\sin\!\bigl(
k_i\mathbf d_i^\top[x,y]^\top{}-\omega_i t+\phi_i\bigr),
\end{aligned}
\label{eq:supp_wave_field}
\end{equation}
where $A_i$, $k_i$, $\omega_i$, $\phi_i$, and $\mathbf d_i$ denote the simulation-space amplitude, wave number, angular frequency, random initial phase, and propagation-direction unit vector, respectively. The dominant wave direction is randomized for each episode, and the eight components are distributed around it. Given reference-scale wave parameters, the ship normalization coefficient is applied as
\begin{equation}
\begin{aligned}
A_i^{\mathrm{sim}}&=\eta_L A_i^{\mathrm{ref}},
&H_s^{\mathrm{sim}}&=\eta_L H_s^{\mathrm{ref}},\\
\lambda_i^{\mathrm{sim}}&=\eta_L\lambda_i^{\mathrm{ref}},
&k_i^{\mathrm{sim}}&=\frac{k_i^{\mathrm{ref}}}{\eta_L}.
\end{aligned}
\label{eq:supp_wave_scaling}
\end{equation}
We correct the component frequencies through an equivalent-gravity parameter. Under the deep-water dispersion relation used to construct the wave components, we set
\begin{equation}
g_{\mathrm{eq}}=\eta_L g,
\qquad
\omega_i^{\mathrm{sim}}
=\sqrt{g_{\mathrm{eq}}k_i^{\mathrm{sim}}}
=\sqrt{gk_i^{\mathrm{ref}}}
=\omega_i^{\mathrm{ref}}.
\label{eq:supp_equivalent_gravity}
\end{equation}
Thus, spatial quantities are scaled by $\eta_L$ while the reference time scale is retained, $t^{\mathrm{sim}}=t^{\mathrm{ref}}$. Equivalently, the wave phase $\omega_i t$ and the ratios $H_s/L$ and $\lambda_i/L$ are preserved. The equivalent gravity is used for wave-frequency construction; it does not introduce an additional force term into the UAV controller.

The reference-scale significant-wave-height and wind-speed ranges are listed in Table~\ref{tab:supp_sea_states}; their runtime simulation values are obtained using Eqs.~\eqref{eq:supp_wave_scaling} and~\eqref{eq:supp_uav_scaling}. Sea-surface occlusion is approximated by height comparison: a camera-to-target ray is marked as occluded if the sampled water-surface height exceeds the ray height at any sampled point. This avoids repeated expensive mesh intersections with the dynamic surface.

\begin{table}[t]
\centering
\small
\begin{tabular}{ccc}
\toprule
State & $H_s$ (m) & $U_{\mathrm{ref}}$ (m/s) \\
\midrule
0 & 0 & 0--0.2 \\
1 & 0--0.1 & 0.3--1.5 \\
2 & 0.1--0.5 & 1.6--3.3 \\
3 & 0.5--1.25 & 3.4--5.4 \\
4 & 1.25--2.5 & 5.5--7.9 \\
5 & 2.5--4 & 8.0--10.7 \\
6 & 4--6 & 10.8--13.8 \\
7 & 6--9 & 13.9--17.1 \\
8 & 9--14 & 17.2--20.7 \\
9 & 14--20 & 20.8--24.4 \\
\bottomrule
\end{tabular}
\caption{Reference-scale sea-state parameter ranges before ship-specific normalization. At runtime, $H_s$ and $U_{\mathrm{ref}}$ are multiplied by $\eta_L$.}
\label{tab:supp_sea_states}
\end{table}

\subsection{Wave-induced Vessel Motion}
We define bow, stern, port, and starboard anchor points on the waterline, with surface heights $h_{\mathrm{bow}}$, $h_{\mathrm{stern}}$, $h_{\mathrm{port}}$, and $h_{\mathrm{starboard}}$. For vessel length $L$ and beam $B$, the simplified rigid-body response is
\begin{equation}
z_{\mathrm{heave}}=\frac{h_{\mathrm{bow}}+h_{\mathrm{stern}}+h_{\mathrm{port}}+h_{\mathrm{starboard}}}{4},
\end{equation}
\begin{equation}
\theta_{\mathrm{pitch}}=\arctan\!\frac{h_{\mathrm{bow}}-h_{\mathrm{stern}}}{L},
\end{equation}
\begin{equation}
\theta_{\mathrm{roll}}=\arctan\!\frac{h_{\mathrm{port}}-h_{\mathrm{starboard}}}{B}.
\end{equation}
The vessel's yaw relative to the world frame is randomized at reset. Thereafter, the vessel undergoes wave-driven translation, roll, and pitch, yielding time-varying motion that perturbs reconstruction without introducing an unrelated high-fidelity six-degree-of-freedom hydrodynamic model.

\subsection{Wind Field and Effective Flight Time}
Wind direction is randomized relative to the dominant wave direction. We also model vertical wind shear: wind speed is lower near the sea surface because of friction and increases with altitude toward the episode-level reference speed. To remain consistent with the unchanged time scale in Eq.~\eqref{eq:supp_equivalent_gravity}, UAV and wind velocities are scaled by the same normalization coefficient as spatial distances:
\begin{equation}
\begin{aligned}
\mathbf W^{\mathrm{sim}}&=\eta_L\mathbf W^{\mathrm{ref}},\\
V_a^{\mathrm{sim}}&=\eta_L V_a^{\mathrm{ref}},\\
v_g^{\mathrm{sim}}&=\eta_L v_g^{\mathrm{ref}}.
\end{aligned}
\label{eq:supp_uav_scaling}
\end{equation}
Let $\mathbf{W}$ be the scaled local wind vector, $\mathbf{u}$ the unit vector from the UAV to the target viewpoint, and $V_a$ the scaled static-air speed. Decomposing the wind into parallel and lateral components, $\mathbf{W}_{\parallel}=(\mathbf{W}^\top\mathbf{u})\mathbf{u}$ and $\mathbf{W}_{\perp}=\mathbf{W}-\mathbf{W}_{\parallel}$, the effective ground speed along the desired track is
\begin{equation}
v_g=\mathbf{W}^\top\mathbf{u}+
\sqrt{V_a^2-\lVert\mathbf{W}-(\mathbf{W}^\top\mathbf{u})\mathbf{u}\rVert_2^2}.
\label{eq:supp_ground_speed}
\end{equation}
For a voxel $v$ with center $x_v$ and a candidate position $q$, we use the Euclidean straight-line distance
\begin{equation}
\delta_t(v,q)=\lVert q-x_v\rVert_2
\label{eq:supp_candidate_distance}
\end{equation}
as the distance feature. This approximation is inexpensive, is consistent with the local action parameterization, and provides the scorer with a smooth measure of the travel and viewing-scale trade-off without introducing an additional path planner inside the candidate evaluator.

For a reference-scale target distance $d^{\mathrm{ref}}$, the runtime distance is $d^{\mathrm{sim}}=\eta_Ld^{\mathrm{ref}}$. Because distance and velocity use the same coefficient, the corresponding flight duration satisfies
\begin{equation}
\Delta t^{\mathrm{sim}}
=\frac{d^{\mathrm{sim}}}{v_g^{\mathrm{sim}}}
=\frac{d^{\mathrm{ref}}}{v_g^{\mathrm{ref}}}
=\Delta t^{\mathrm{ref}}.
\label{eq:supp_time_scaling}
\end{equation}
During this interval, the wave field and ship pose continue to evolve. Thus, headwind increases the time before the next observation and indirectly increases the magnitude of unmodeled ship motion between consecutive frames; tailwind has the opposite effect. We use the reference-scale airspeed $V_a^{\mathrm{ref}}=25$ m/s and set the runtime value to $V_a^{\mathrm{sim}}=25\eta_L$ m/s.

\subsection{Initial State and Camera}
At reset, the UAV is sampled from the upper hemisphere at a reference-scale distance of 5--10 m from the ship center, corresponding to a runtime distance of $5\eta_L$--$10\eta_L$ m, and approximately faces the center with a random angular perturbation of $\pm30^\circ$. The simulated camera resolution is $400\times400$ pixels, and its vertical field of view is $90^\circ$.

\end{document}